\documentclass[letterpaper]{article} 
\usepackage[preprint]{aaai2027}  
\usepackage[hyphens]{url}  
\usepackage{graphicx} 
\usepackage{natbib}  
\usepackage{caption} 
\usepackage{amsmath}
\usepackage{algorithm}
\usepackage{algpseudocode}
\usepackage{subcaption}
\usepackage{amsmath}
\usepackage{booktabs}
\usepackage{multirow}
\usepackage{algpseudocode}
\usepackage{listings}
\usepackage{newfloat}
\usepackage{listings}
\DeclareCaptionStyle{ruled}{labelfont=normalfont,labelsep=colon,strut=off} 
\floatstyle{ruled}
\newfloat{listing}{tb}{lst}{}
\floatname{listing}{Listing}

\usepackage{booktabs}

\title{Mimir: A Neuro-Symbolic Memory System with Dynamic Grounding for Embodied Agents in Interactive Environments}
\author{
    Haoming Xu\textsuperscript{\rm 1,\rm 3}\equalcontrib,
    Zhenlin He\textsuperscript{\rm 1,\rm 2}\equalcontrib,
    Hengyi Wang\textsuperscript{\rm 1,\rm 2},
    Jiafeng Xu\textsuperscript{\rm 2},
    Hao Dong\textsuperscript{\rm 1, \rm 2}
}

\affiliations{
    \textsuperscript{\rm 1}PrimeBot Research Institute\\
    \textsuperscript{\rm 2}Peking University\\
    \textsuperscript{\rm 3}
    University of Chinese Academy of Sciences\\
    
}

\begin{document}

\maketitle

\begin{abstract}
Long-horizon embodied task requires agents to act under partial observability while preserving both scene belief and execution progress. Flat histories or implicit policy states may contain past observations, but they do not provide an explicit interface for deciding which world facts support the currently active goal. We introduce \textsc{Mimir}, a neuro-symbolic memory that separates world memory from task memory and dynamically grounds them before each action.
World memory maintains object locations, object states, and perceptual evidence, while task memory maintains an ordered goal agenda, progress state, hand state, failures, and execution constraints.
A grounding module binds the active goal to recalled world candidates, fills missing source locations, and attaches evidence before planning and embodiment-specific execution.
Across tested backbones, \textsc{Mimir} consistently improves on different EB-ALFRED and EB-Habitat tasks, with maximum gains of 42.5\% and average gains of 23.0\%, respectively. Compared with the best results among prior agent and memory systems evaluated under the same backbone, \textsc{Mimir} improves the overall average success rate by 8.5\%. Finally, on the EB-Habitat Long-horizon subset, \textsc{Mimir} achieves 86.0\% success rate, substantially outperforming current closed-source models. Our code will be released soon.

\end{abstract}

\section{Introduction}
\begin{figure*}
    \centering
    \includegraphics[width=\linewidth]{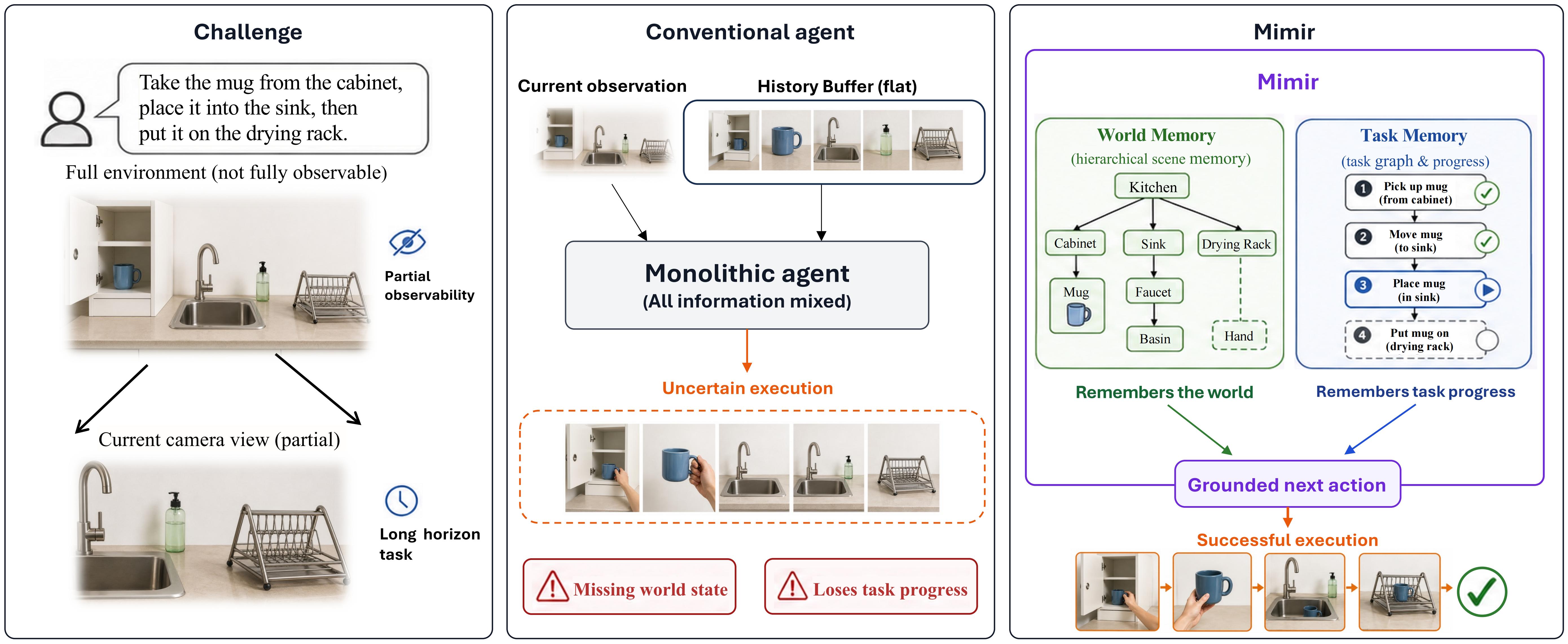}
    \caption{Motivation of \textsc{Mimir}.
Long-horizon embodied tasks require memory of both physical world state and task progress.
Flat-history agents can retain past observations yet still lose object locations or confuse completed and pending subgoals.
\textsc{Mimir} maintains explicit world and task memories, and dynamically grounds the active goal against recalled world evidence before action.}
    \label{fig:stage}
\end{figure*}
Long-horizon embodied manipulation operates under partial observability. As an agent explores and alters a scene, relevant objects may leave view, change state, or become identifiable only through earlier interactions. Current observations are often insufficient for selecting the next action. The agent must preserve the evolving world and its task progress.

The challenge is not merely to retain past frames. In Fig.~\ref{fig:stage}, a flat history may retain evidence without exposing an actionable state: whether an object remains in its receptacle, is held by the agent, or has been delivered, and whether a subgoal is complete, pending, or blocked. Effective memory must evolve with the world and support the next decision.

Recent embodied agents have advanced instruction decomposition, skill selection, and multimodal control by combining language, vision, and action \cite{ahn2022saycan,song2023llmplanner,driess2023palme,brohan2023rt2}. Yet they must maintain state over time. Hidden objects require persistent world state, while multi-step execution requires explicit task state. Long prompts or implicit policy states may preserve information, but provide no interface for inspecting, updating, and reusing it.

Memory-based systems make parts of this state explicit. LLM-State tracks object attributes and changes, while CAPEAM records action context and altered object arrangements for planning \cite{chen2023llmstate,kim2023capeam}. Both strengthen world-state tracking, but leave task progress implicit and do not bind recalled evidence to the active goal. CLEA integrates environmental memory with execution critique and replanning, whereas RoboMemory coordinates multiple cognitive memory types \cite{lei2025clea,lei2025robomemory}. However, neither provides a direct interface between world belief and task progress.

Other methods address complementary functions. AutoManual and MSI-Agent convert experience into reusable manuals or insights \cite{chen2024automanual,fu2024msiagent}, while MemoryVLA and related methods inject temporal context into policy execution \cite{memoryvla2025,memer2025,mapvla2025}. They emphasize reusable knowledge or architecture-specific policy memory rather than a transparent episode-level decision state.

These limitations expose a missing operational link between what an agent knows and what it must accomplish next. We propose \textsc{Mimir}, a neuro-symbolic memory system that separates world memory from task memory and grounds the active goal against recalled world evidence before each action. World memory stores object locations, states, and perceptual evidence, while task memory tracks the ordered goal agenda, execution progress, hand state, failures, and constraints. By binding the active goal to its object, source, target, and supporting evidence, \textsc{Mimir} turns memory from a passive history into an explicit online decision state. Our contributions are threefold.

\begin{itemize}
    \item We formulate embodied memory as the interaction between world state and task state, separating scene belief from execution progress and identifying dynamic grounding as the key operation before action.

    \item We instantiate this formulation in \textsc{Mimir}, a closed-loop memory system that maintains an ordered task memory and an episode-level TreeMemory of object locations, states, and evidence. It grounds each active goal through memory recall, tracks hand state and execution failures, and translates grounded abstract actions through embodiment-specific adapters.

    \item Extensive evaluations across different multimodal backbones of EB-ALFRED and EB-Habitat show improvements in both success rate and goal completion for thirteen backbones, with maximum and mean success-rate gains of 42.5\% and 23.0\%. Under a matched-backbone comparison, \textsc{Mimir} improves the overall average success rate over prior agent and memory systems by 8.5\% and reaches 86.0\% on EB-Habitat Long-horizon, substantially above current closed-source models.
\end{itemize}

\section{Related Work}

\subsection{Memory in Large Language Model Agents.}
Memory is a core mechanism for reasoning beyond the current context. Cognitive accounts distinguish working, episodic, and semantic memory \cite{atkinson1968human,baddeley1974working,tulving1972episodic}, and recent language agents instantiate these ideas by storing interaction histories, reflections, skills, or structured user memories for later reuse \cite{yao2023react,shinn2023reflexion,wang2023voyager,packer2023memgpt,chhikara2025mem0}. These memories improve adaptation across trials and long-horizon interaction, but they remain mostly textual or symbolic; embodied manipulation requires memory grounded in physical state, object relations, and task progress.

\subsection{Memory for Embodied Planning}
LLM and VLM-based embodied planners decompose goals, ground actions, and select executable skills \cite{ahn2022saycan,song2023llmplanner,driess2023palme,zhang2023coela,lei2025clea}, and later methods add manuals, experience stores, symbolic knowledge, or structured scene memory to guide decision-making across interactions \cite{chen2024automanual,fu2024msiagent,glocker2025llm,choi2025nesyc,karma2024}. RoboMemory organizes spatial, temporal, episodic, and semantic memory in a brain-inspired architecture for physical embodied systems \cite{lei2025robomemory}, with follow-up work refining evolving or compiled memory and KV-cache-centric memory management \cite{ma2026brainmem,memcompiler2026,yang2026keep}.

LLM-State maintains an open-world representation of object attributes and their changes for retrospective long-horizon planning \cite{chen2023llmstate}; \textsc{Mimir} instead separates the external world belief from an execution-progress agenda and binds only the active goal to recalled evidence before each action.
CAPEAM conditions subgoal planning on prior action context and stores changed object arrangements and states for navigation and interaction \cite{kim2023capeam}; \textsc{Mimir} makes the update interface explicit through distinct observation and action writes and carries the selected memory evidence into a grounded goal.
SayPlan searches hierarchical 3D scene graphs and iteratively replans with scene-graph-simulator feedback \cite{rana2023sayplan}; \textsc{Mimir} does not assume a 3D scene graph and instead serializes the current first-person observation, ordered task state, recalled candidates, failures, and world-memory snapshot into the VLM decision input.

A complementary line of work models the environment itself: open-vocabulary maps, multimodal 3D maps, and scene graphs make environments queryable for navigation and spatial reasoning \cite{huang2022vlmaps,jatavallabhula2023conceptfusion,gu2024conceptgraphs,chang2023goat,loo2025openscenegraphs}, and recent spatial-memory systems maintain persistent 3D or semantic-spatial representations for exploration and embodied reasoning \cite{yang20253dmem,3dllmmem2025,metamemory2025,zhang2025nava3}. However, world memory alone does not encode the task history of the agent; existing work has not fully unified world-level and task-level memory.

\subsection{Memory in Vision-Language-Action Policies.}
Vision-language-action policies map language and visual observations directly to robot actions \cite{brohan2023rt2,oneill2023openxembodiment,kim2024openvla,black2024pi0}. To address the non-Markovian nature of long-horizon manipulation, recent VLA models integrate perceptual-cognitive memory, retrieve prior experience and keyframes, or use demonstration-derived memory prompts and episodic memory during policy execution \cite{memoryvla2025,memer2025,mapvla2025,echovla2025,chameleon2026}, with further variants exploring multi-scale, recurrent, and test-time physical memory \cite{mem2026,rememvla2026,helm2026,vpwem2026,physmem2026}. These approaches are tightly coupled to model architecture and action generation, providing less explicit organization for general embodied memory across planning and execution.
\begin{figure*}[t]
    \centering
    \includegraphics[width=\textwidth]{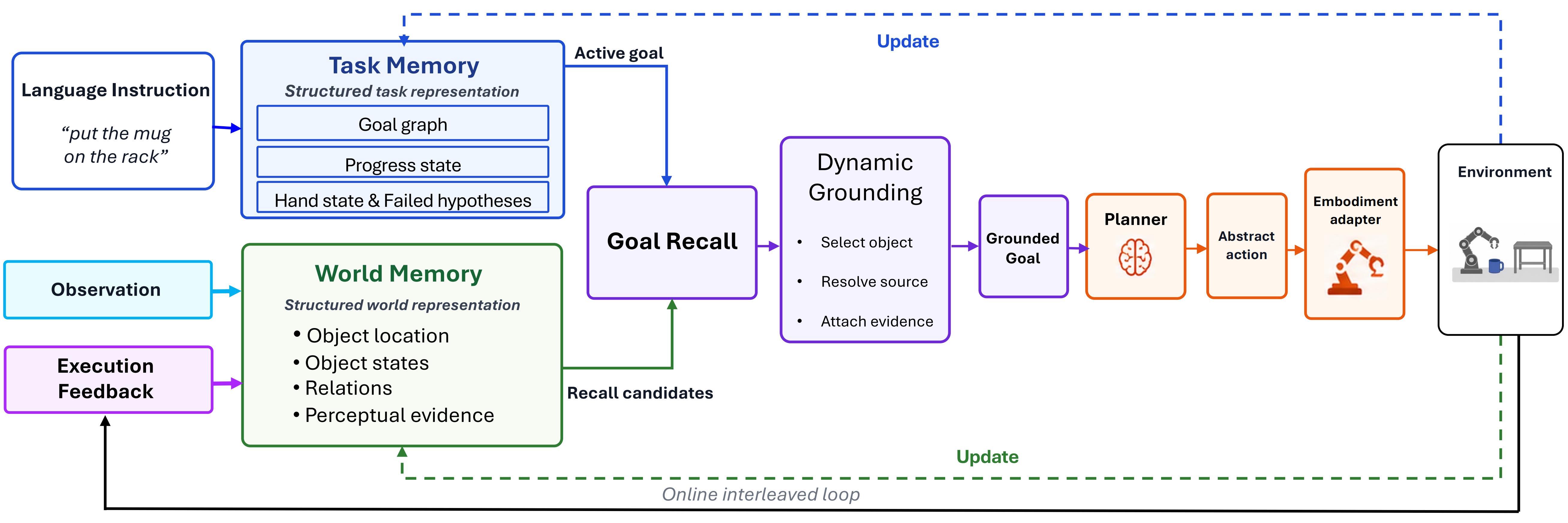}
    \caption{Overview of \textsc{Mimir}.
Task memory preserves execution
progress, while world memory maintains scene entities, relations, and
evidence. Goal-conditioned recall retrieves candidates for the active
goal, and dynamic grounding selects an evidence-supported binding for
planning and embodiment-specific execution. Observations and execution
feedback update the two memories online.}
    \label{fig:structure}
\end{figure*}

\section{Method}
\label{sec:method}

We present \textsc{Mimir}, a neuro-symbolic memory system for
long-horizon embodied manipulation. At the beginning of an episode, the
agent receives an instruction $x$ and the legal action space $A_e$ of
embodiment $e$. At step $t$, it observes $o_t$ and, after executing a
legal action $a_t\in A_e$, receives feedback $y_t$. \textsc{Mimir}
maintains
\begin{equation}
M_t=(T_t,W_t),
\label{eq:mimir_memory}
\end{equation}
where task memory $T_t$ records execution progress and world memory
$W_t$ records the current scene belief.

The two memories are separated because they evolve from different
evidence. Observations and interaction effects revise object locations
and states in world memory, whereas execution outcomes revise the goal
agenda, hand state, and failed hypotheses in task memory. A flat history
leaves both states implicit and requires the planner to reconstruct them
at every step. \textsc{Mimir} instead preserves them as persistent and
independently updateable memories.

Dynamic grounding connects the two memories. The current observation
first updates world memory, after which task memory exposes the next
unfinished goal. Goal-conditioned recall retrieves a bounded set of
relevant entities and evidence, and grounding binds the goal to a
concrete object and its source and target context. The grounded goal is
planned in an embodiment-independent action space and mapped to $A_e$;
execution feedback then updates both memories. The same loop supports
separate exploration before execution or interleaved exploration when
the current memory cannot support a grounded action.

\subsection{Task Memory}
\label{sec:task_memory}

Task memory represents what remains to be accomplished and what
execution has already established. It comprises symbolic goals
$\mathcal G$, an agenda order $\preceq$, a goal-status map $s_t$, the
current hand state $h_t$, failed object--source hypotheses $F_t$, and
evidence $C_t$ supporting completed goals:
\begin{equation}
T_t=(\mathcal G,\preceq,s_t,h_t,F_t,C_t).
\label{eq:task_memory}
\end{equation}
The map $s_t$ assigns each goal one of three states: pending, completed,
or blocked. The order $\preceq$ places prerequisites before dependent
goals and uses task order as a deterministic tie breaker. Task memory
therefore preserves both the remaining agenda and the execution state
needed to interpret it. In contrast to a history buffer, these fields are
updated independently: completing one goal does not remove scene
evidence, and revising a world hypothesis does not erase established task
progress. This separation gives grounding a stable task-side query even
when the world representation continues to change.

For $g\in\mathcal G$, let $\mathcal P_t$ denote the pending-goal set and
$g_t$ the active goal. Here $\bot_{\mathrm{goal}}$ denotes the absence of
an active goal, and $\min_{\preceq}$ selects the first goal under the
agenda order:
\begin{equation}
\begin{aligned}
\mathcal P_t
&=\{g\in\mathcal G\mid s_t(g)=\mathrm{pending}\},\\
g_t
&=
\begin{cases}
\min_{\preceq}\mathcal P_t,
    & \mathcal P_t\neq\emptyset,\\
\bot_{\mathrm{goal}},
    & \mathcal P_t=\emptyset.
\end{cases}
\end{aligned}
\label{eq:goal_selection}
\end{equation}
Because $\mathcal P_t$ is recomputed after every task-memory update, the
next goal follows from recorded execution state rather than a fresh
interpretation of the full history. Completed goals are excluded from
later planning, blocked goals remain explicit, and a dependent goal is
not selected before its prerequisites. Task memory is therefore not an
action transcript: it retains unfinished goals, the held object, failed
bindings, and evidence for completed work, preventing repeated failures
or redundant execution.

We represent environment feedback as
\begin{equation}
y_t=(\sigma_t,\Delta_t),
\label{eq:execution_feedback}
\end{equation}
where $\sigma_t$ is the action-level outcome and $\Delta_t$ contains any
feedback-supported change in object relation, object state, or hand
state. The effect may be empty when an action succeeds without
establishing a task-relevant change. Given the active goal, executed
action, and feedback, task memory is updated by
\begin{equation}
T_{t+1}=U_T(T_t,g_t,a_t,y_t).
\label{eq:task_update}
\end{equation}

Action success does not directly imply completion. The active goal is
completed only when the supported effect establishes its postcondition:
\begin{equation}
s_{t+1}(g_t)=\mathrm{completed}
\quad\Longleftrightarrow\quad
\operatorname{Post}(g_t,\Delta_t)=1.
\label{eq:goal_completion}
\end{equation}
Navigation or opening may therefore succeed while a manipulation goal
remains pending; pick, placement, and state-change goals require the
corresponding hand, relation, or state effect. Supported grasp, place, or
release effects update $h_t$, a contradicted object--source binding enters
$F_{t+1}$, and verified completion evidence enters $C_{t+1}$.

\subsection{World Memory}
\label{sec:world_memory}

World memory is an episode-level structured memory of scene entities
and their evolving relations. At step $t$, it comprises an entity set
$V_t$, parent relations $E_t$, a scene root $r$, a label function
$\ell_t$, and an attribute-and-evidence store $\phi_t$:
\begin{equation}
W_t=(V_t,E_t,r,\ell_t,\phi_t).
\label{eq:world_memory}
\end{equation}
For each non-root entity $v\in V_t$, the relations induce an active
parent $p_t(v)$ representing its remembered room, supporting surface,
container, or held state. The function $\ell_t$ records semantic
identity, while $\phi_t$ stores observed attributes, interaction-supported
state, and supporting evidence. The evidence store preserves the
observation or interaction from which a record was obtained, allowing a
later decision to distinguish an observed relation from one supported by
execution feedback. Each entity has one active parent at a time, while
uncertain alternatives are represented as separate hypotheses.

This representation places spatial and manipulation state in a common
memory. An object may initially belong to a receptacle, later be
associated with the hand after a successful pick, and finally be
associated with a target after placement. Because these records persist
after an entity leaves the current view, the agent can reason about
hidden objects without replaying the visual history.

Observation and interaction update world memory through distinct
operators:
\begin{equation}
\begin{aligned}
W_t^{+}
&=U_W^{\mathrm{obs}}(W_t,o_t),\\
W_{t+1}
&=U_W^{\mathrm{act}}(W_t^{+},a_t,y_t).
\end{aligned}
\label{eq:world_update}
\end{equation}
The observation update adds new entities or refines a hypothesis when
identity and relational context agree with stored evidence. Additional
same-name instances or incompatible relations remain separate
hypotheses, and unresolved alternatives are retained for grounding. The
action update writes only effects supported by $\Delta_t$: a pick may
associate an object with the hand, a placement with the target, and a
state-changing action with a revised state. Failed actions leave physical
relations unchanged, while contradicted bindings are recorded in task
memory. Together, the two updates preserve both current perception and
interaction-supported changes that may later leave view.

For the active goal $g_t$, the query constructor produces
$q_t=Q(g_t)$ from its object description, required state, and source or
target relations. The retrieval operator searches $W_t^{+}$ and returns
an ordered and bounded list
\begin{equation}
\mathcal R_t
=R(W_t^{+},q_t)=
\left[
\bigl(v_k,p_t^{+}(v_k),\eta_{t,k}\bigr)
\right]_{k=1}^{K_t},
\label{eq:world_recall}
\end{equation}
where $v_k$ is a candidate entity, $p_t^{+}(v_k)$ its remembered parent,
and $\eta_{t,k}$ the lexical, visual, semantic, and relational evidence
supporting it. The candidate pool contains entities whose identity,
attributes, or relations are compatible with the active query. Evidence
may therefore support a candidate through its name, appearance, category,
remembered parent, or relation to another entity. Returning a bounded
list creates a compact, goal-conditioned view of world memory rather than
exposing the full accumulated scene record. The order retains the
relative support among candidates, while bindings already contradicted in
$F_t$ are excluded from repeated selection.

\subsection{Grounding and Execution}
\label{sec:grounding_execution}

Recall identifies which world-memory hypotheses may support the active
goal; grounding determines which one should instantiate it at the
current step. The operator $G$ first removes candidates that cannot
support a legal interaction, conflict with explicit source or target
constraints, or have been contradicted by feedback. It then selects the
candidate best supported by the current observation and recalled
evidence. Finally, it resolves an omitted source from the selected
entity's remembered parent and attaches the supporting evidence to the
grounded goal. Recall and grounding therefore play different roles:
recall preserves multiple plausible hypotheses, whereas grounding commits
the current decision to one admissible, evidence-supported binding.

Let $u_x(g_t)$ denote an explicitly specified source and
$\bot_{\mathrm{src}}$ the absence of one. For selected candidate
$k^\star$, let $v_t^\star=v_{k^\star}$ and
$\eta_t^\star=\eta_{t,k^\star}$. The target $d_t^\star$ is retained from
the active goal, while the source is
\begin{equation}
\begin{aligned}
u_t^\star
&=
\begin{cases}
u_x(g_t),
    & u_x(g_t)\neq\bot_{\mathrm{src}},\\
p_t^{+}(v_t^\star),
    & u_x(g_t)=\bot_{\mathrm{src}},
\end{cases}\\
\hat g_t
&=G(g_t,\mathcal R_t,F_t,A_e,x,o_t)\\
&=(g_t,v_t^\star,u_t^\star,d_t^\star,\eta_t^\star).
\end{aligned}
\label{eq:grounded_goal}
\end{equation}
The tuple makes the selected world evidence explicit to the planner. An
explicit source is preserved, whereas an omitted source can be completed
from world memory. The target remains a task requirement rather than a
newly inferred destination. If no candidate satisfies the task and
action constraints, $G$ returns $\bot_{\mathrm{grd}}$ rather than
constructing an unsupported binding. This fail-closed interface prevents
an uncertain memory query from being converted directly into an
executable action.

When several candidates remain, the VLM compares the current observation
with their compact memory descriptions, conditioned on $x$, $g_t$, and
relevant failures in $F_t$. Its output is restricted to
$\{v_k\}_{k=1}^{K_t}$, allowing visual disambiguation without introducing
an entity or location unsupported by memory.

Let $\bar{\mathcal A}$ denote the embodiment-independent abstract action
space. The planner produces $\bar a_t\in\bar{\mathcal A}$, and the
adapter $\rho_e$ maps it to a legal action or returns
$\bot_{\mathrm{act}}$:
\begin{equation}
\begin{aligned}
\bar a_t
&=\Pi(\hat g_t,T_t,W_t^{+}),\\
\rho_e
&:\bar{\mathcal A}\rightarrow A_e\cup\{\bot_{\mathrm{act}}\},\\
a_t
&=\rho_e(\bar a_t).
\end{aligned}
\label{eq:action_mapping}
\end{equation}

Let $\operatorname{Step}_e$ denote the environment transition for
embodiment $e$. Given a legal action $a_t\in A_e$, it returns the next
observation $o_{t+1}$, execution feedback $y_t$, and a termination flag
$z_{t+1}$. Here $z_t=0$ denotes an active episode and $z_t=1$ denotes
termination.

Algorithm~\ref{alg:mimir_loop} summarizes the action bearing memory
loop. The environment transition is invoked only when grounding and
action mapping produce a valid result. Otherwise, the system follows
the configured exploration or termination policy without passing an
unsupported value to the environment.
\begin{algorithm}[t]
\caption{\textsc{Mimir} Closed-Loop Inference}
\label{alg:mimir_loop}
\begin{algorithmic}[1]
\Require $x,T_0,W_0,A_e,o_0$
\State $z_0\gets 0,\quad t\gets 0$

\While{$z_t=0\land\mathcal P_t\neq\emptyset$}
    \State $W_t^{+}\gets U_W^{\mathrm{obs}}(W_t,o_t)$
    \State $g_t\gets\min_{\preceq}\mathcal P_t$

    \State $\mathcal R_t\gets R(W_t^{+},Q(g_t))$
    \State $\hat g_t\gets
        G(g_t,\mathcal R_t,F_t,A_e,x,o_t)$

    \If{$\hat g_t=\bot_{\mathrm{grd}}$}
        \State $a_t\gets
            \operatorname{Explore}_e(g_t,T_t,W_t^{+},A_e)$
    \Else
        \State $\bar a_t\gets\Pi(\hat g_t,T_t,W_t^{+})$
        \State $a_t\gets\rho_e(\bar a_t)$

        \If{$a_t=\bot_{\mathrm{act}}$}
            \State $\bar a_t\gets
                \operatorname{Replan}
                (\hat g_t,T_t,W_t^{+},\bar a_t)$
            \State $a_t\gets\rho_e(\bar a_t)$
        \EndIf
    \EndIf

    \If{$a_t=\bot_{\mathrm{act}}$}
        \State $T_t\gets\operatorname{Block}(T_t,g_t)$
        \State \textbf{continue}
    \EndIf

    \State $(o_{t+1},y_t,z_{t+1})
        \gets\operatorname{Step}_e(a_t)$
    \State $W_{t+1}\gets
        U_W^{\mathrm{act}}(W_t^{+},a_t,y_t)$
    \State $T_{t+1}\gets
        U_T(T_t,g_t,a_t,y_t)$
    \State $t\gets t+1$
\EndWhile

\State \Return $(M_t,\operatorname{Status}(T_t,z_t))$
\end{algorithmic}
\end{algorithm}

Execution feedback closes the loop through
Equations~\ref{eq:task_update} and~\ref{eq:world_update}. World memory
records supported changes in object relations and states, while task
memory determines whether these changes complete the active goal.
Intermediate actions may therefore update the world or hand state while
the goal remains pending. Contradicted bindings are retained as failure
evidence and excluded from later grounding.

\section{Experiments}
To evaluate Mimir’s long-horizon task-solving capability, we follow prior embodied-agent evaluations and conduct experiments on EB-ALFRED and EB-Habitat from EmbodiedBench (Yang et al. 2025). Our experiments demonstrate three key findings: 
\begin{itemize}
    \item Mimir consistently improves task-planning performance across multimodal large language models of different scales.
    
    \item Under the same backbone, Mimir achieves higher success rates than existing methods.
    
    \item With a mid-sized open-source backbone, Mimir substantially outperforms state-of-the-art closed-source models.
\end{itemize}

\subsection{Metrics}
\label{sec:experiments_metrics}
We report Success Rate (SR) and Goal Condition Success Rate (GC), following EmbodiedBench and prior memory-augmented embodied-agent evaluations \cite{yang2025embodiedbench,lei2025robomemory}. SR measures full-task completion, while GC measures completion of ordered goal conditions. Following the setup of prior work, We set temperature = 0 to avoid randomness during the experiment which aligning with the EmbodiedBench setting. Each task is executed once, And we repeated the experiments multiple times and consistently obtained the same results.

\begin{table*}[h]
\centering
\small
\setlength{\tabcolsep}{10pt}
\renewcommand{\arraystretch}{0.9}
\begin{tabular}{@{}lrr@{\hspace{7pt}}rr@{\hspace{16pt}}rr@{\hspace{7pt}}rr@{}}
\toprule
Backbone
& \multicolumn{4}{c}{EB-ALFRED}
& \multicolumn{4}{c}{EB-Habitat} \\
\cmidrule(lr){2-5}\cmidrule(lr){6-9}
& \multicolumn{1}{c}{SR}
& \multicolumn{1}{c}{GC}
& \multicolumn{1}{c}{$\Delta$SR}
& \multicolumn{1}{c}{$\Delta$GC}
& \multicolumn{1}{c}{SR}
& \multicolumn{1}{c}{GC}
& \multicolumn{1}{c}{$\Delta$SR}
& \multicolumn{1}{c}{$\Delta$GC} \\
\midrule
Qwen3-VL-8B
& 26.5 & 39.7 & &
& 28.5 & 37.1 & & \\
\hspace{1em}w/ \textsc{Mimir}
& \textbf{57.0} & \textbf{66.7}
& \textbf{+30.5} & \textbf{+26.9}
& \textbf{65.0} & \textbf{73.8}
& \textbf{+36.5} & \textbf{+36.8} \\
\addlinespace[1.0pt]

Cosmos-Reason2-8B
& 13.0 & 21.8 & &
& 28.5 & 37.4 & & \\
\hspace{1em}w/ \textsc{Mimir}
& \textbf{51.5} & \textbf{63.4}
& \textbf{+38.5} & \textbf{+41.6}
& \textbf{45.5} & \textbf{61.1}
& \textbf{+17.0} & \textbf{+23.6} \\
\addlinespace[1.0pt]

InternVL3-8B
& 17.5 & 28.6 & &
& 26.5 & 37.3 & & \\
\hspace{1em}w/ \textsc{Mimir}
& \textbf{60.0} & \textbf{68.5}
& \textbf{+42.5} & \textbf{+39.9}
& \textbf{60.5} & \textbf{69.2}
& \textbf{+34.0} & \textbf{+31.9} \\
\addlinespace[1.0pt]

Qwen3.5-8B
& 41.5 & 52.0 & &
& 32.0 & 38.3 & & \\
\hspace{1em}w/ \textsc{Mimir}
& \textbf{59.0} & \textbf{67.6}
& \textbf{+17.5} & \textbf{+15.6}
& \textbf{53.0} & \textbf{66.2}
& \textbf{+21.0} & \textbf{+27.9} \\
\addlinespace[1.0pt]

InternVL3.5-8B
& 24.5 & 34.1 & &
& 37.0 & 45.1 & & \\
\hspace{1em}w/ \textsc{Mimir}
& \textbf{60.5} & \textbf{69.5}
& \textbf{+36.0} & \textbf{+35.4}
& \textbf{49.5} & \textbf{62.9}
& \textbf{+12.5} & \textbf{+17.7} \\
\addlinespace[1.0pt]

Gemma-3-12B
& 29.0 & 41.4 & &
& 26.5 & 33.0 & & \\
\hspace{1em}w/ \textsc{Mimir}
& \textbf{59.0} & \textbf{68.0}
& \textbf{+30.0} & \textbf{+26.6}
& \textbf{64.0} & \textbf{75.7}
& \textbf{+37.5} & \textbf{+42.6} \\
\addlinespace[1.0pt]

Gemma-3-27B
& 35.0 & 46.2 & &
& 42.5 & 49.7 & & \\
\hspace{1em}w/ \textsc{Mimir}
& \textbf{59.0} & \textbf{67.7}
& \textbf{+24.0} & \textbf{+21.5}
& \textbf{60.0} & \textbf{71.3}
& \textbf{+17.5} & \textbf{+21.6} \\
\addlinespace[1.0pt]

Qwen3.5-27B
& 70.5 & 75.1 & &
& 57.0 & 64.4 & & \\
\hspace{1em}w/ \textsc{Mimir}
& \textbf{72.5} & \textbf{79.5}
& \textbf{+2.0} & \textbf{+4.4}
& \textbf{62.5} & \textbf{71.5}
& \textbf{+5.5} & \textbf{+7.1} \\
\addlinespace[1.0pt]

Qwen3-VL-32B
& 50.5 & 59.5 & &
& 50.5 & 57.5 & & \\
\hspace{1em}w/ \textsc{Mimir}
& \textbf{68.0} & \textbf{73.0}
& \textbf{+17.5} & \textbf{+13.5}
& \textbf{71.5} & \textbf{78.6}
& \textbf{+21.0} & \textbf{+21.1} \\
\addlinespace[1.0pt]

InternVL3-38B
& 30.0 & 43.0 & &
& 42.5 & 52.6 & & \\
\hspace{1em}w/ \textsc{Mimir}
& \textbf{58.5} & \textbf{67.2}
& \textbf{+28.5} & \textbf{+24.2}
& \textbf{50.0} & \textbf{62.9}
& \textbf{+7.5} & \textbf{+10.3} \\
\addlinespace[1.0pt]

Qwen2.5-VL-72B
& 50.0 & 58.2 & &
& 49.0 & 56.8 & & \\
\hspace{1em}w/ \textsc{Mimir}
& \textbf{66.0} & \textbf{72.9}
& \textbf{+16.0} & \textbf{+14.7}
& \textbf{70.0} & \textbf{77.4}
& \textbf{+21.0} & \textbf{+20.7} \\
\addlinespace[1.0pt]

InternVL3-78B
& 31.5 & 44.1 & &
& 43.5 & 53.4 & & \\
\hspace{1em}w/ \textsc{Mimir}
& \textbf{59.5} & \textbf{68.4}
& \textbf{+28.0} & \textbf{+24.3}
& \textbf{54.5} & \textbf{68.4}
& \textbf{+11.0} & \textbf{+15.0} \\
\bottomrule
\end{tabular}
\caption{Four-task average performance on EB-ALFRED and EB-Habitat. The averages cover Base, Common-sense, Complex-instruction, and Long-horizon. Success Rate (SR) and Goal Condition Success Rate (GC) are reported in percentage points. For each backbone, the first row reports the backbone alone and the second row adds \textsc{Mimir}. $\Delta$SR and $\Delta$GC denote the corresponding benchmark-specific gains from \textsc{Mimir}, computed before rounding.}
\end{table*}

\subsection{Main Results}
\label{tab:backbone_averages}

Table~\ref{tab:backbone_averages} shows broad gains in both SR and GC across model types and scales. The gains are especially large when the bare backbone is weak: InternVL3-8B gains 42.5 SR points on EB-ALFRED, while Gemma-3-12B gains 37.5 SR points and 42.6 GC points on EB-Habitat. Because SR and GC rise together, \textsc{Mimir} does not merely preserve partial progress, it more reliably turns that progress into complete task.

The scale comparison further suggests that explicit state management can compensate for limited backbone capacity. With \textsc{Mimir}, InternVL3-8B reaches an SR of 60.0 and a GC of 68.5 on EB-ALFRED, together with an SR of 60.5 and a GC of 69.2 on EB-Habitat. These results match or exceed those of its 38B and 78B counterparts. Qwen3-VL-32B nevertheless remains stronger than Qwen3-VL-8B, indicating that backbone capacity and structured memory are complementary rather than interchangeable.

\subsection{Comparison with Agent and Memory Systems}

We compare against the EmbodiedBench results reported by RoboMemory under the Qwen2.5-VL-72B backbone \cite{wang2023voyager,shinn2023reflexion,tan2024cradle,tan2025roboos,lei2025robomemory}.

\begin{table}[t]
\centering
\small
\setlength{\tabcolsep}{10pt}
\renewcommand{\arraystretch}{0.9}
\begin{tabular}{@{}lrrrr@{}}
\toprule
Method & \multicolumn{2}{c}{EB-ALFRED} & \multicolumn{2}{c}{EB-Habitat} \\
\cmidrule(lr){2-3}\cmidrule(lr){4-5}
& SR & GC & SR & GC \\
\midrule
Voyager    & 44.0 & 63.7 & 49.0 & 69.0 \\
Reflexion  & 29.0 & 43.5 & 47.5 & 58.6 \\
Cradle     & 43.0 & 54.5 & 46.0 & 59.6 \\
RoboOS     & 22.0 & 28.0 & 29.0 & 38.0 \\
RoboMemory & 67.0 & \textbf{78.4} & 74.0 & 81.0 \\
\midrule
\textsc{Mimir} & \textbf{68.0} & 76.3 & \textbf{90.0} & \textbf{94.6} \\
\bottomrule
\end{tabular}
\caption{Comparison with agent and memory systems under the Qwen2.5-VL-72B. Each value is the unweighted mean of Base and Long-horizon follow RoboMemory.}
\label{tab:memory_systems}
\end{table}

Table~\ref{tab:memory_systems} shows that memory-centered systems substantially outperform generic agent frameworks on both benchmarks. \textsc{Mimir} achieves the highest SR on EB-ALFRED and the strongest SR and GC on EB-Habitat. Compared with RoboMemory, it raises EB-Habitat SR by 16.0 points and GC by 13.6 points. This larger margin on EB-Habitat is consistent with the role of world memory: the planner acts on object and location evidence selected for the active goal, rather than on a general record of prior interaction.

\subsection{Comparison with Foundation Models}

We further compare with recent closed-source multimodal models on EB-Habitat Long-horizon. All entries use a maximum generation budget of 16,384 tokens.

\begin{table}[t]
\centering
\small
\setlength{\tabcolsep}{16pt}
\renewcommand{\arraystretch}{1.04}
\begin{tabular}{@{}lrr@{}}
\toprule
Method & SR & GC \\
\midrule
GPT-5.5 & 78.0 & 79.5 \\
Claude-Sonnet-4-6 & 82.0 & 86.8 \\
Gemini-3.5-Flash & 80.0 & 84.8 \\
Gemini-3.1-Pro & 70.0 & 75.0 \\
Kimi-2.5 & 30.0 & 34.6 \\
\midrule
\textsc{Mimir} (Qwen3-VL-32B) & \textbf{86.0} & \textbf{93.0} \\
\bottomrule
\end{tabular}
\caption{Comparison with closed-source foundation models on EB-Habitat Long-horizon. Success Rate and Goal Condition Success Rate are reported in percentage points.}
\label{tab:foundation_habitat_long}
\end{table}

With a open-source backbone, \textsc{Mimir} reaches an SR of 86.0 and a GC of 93.0, exceeding the strongest closed-source baseline by 4.0 SR points and 6.2 GC points. This result indicates that long-horizon embodied performance depends not only on backbone capacity, but also on how task progress and world evidence are organized for each decision. The structured task--world interface enables a 32B open-source model to outperform leading proprietary models.

\subsection{Ablation Studies}

\begin{table}[t]
\centering
\small
\setlength{\tabcolsep}{4pt}
\renewcommand{\arraystretch}{1.0}
\begin{tabular}{@{}lrrrr@{}}
\toprule
Variant & \multicolumn{2}{c}{EB-ALFRED} & \multicolumn{2}{c}{EB-Habitat} \\
\cmidrule(lr){2-3}\cmidrule(lr){4-5}
& SR & GC & SR & GC \\
\midrule
\multicolumn{5}{l}{\emph{Component removal, Qwen3-VL-8B}} \\
Full \textsc{Mimir} & \textbf{57.0} & \textbf{66.7} & \textbf{65.0} & \textbf{73.8} \\
Without world memory & 52.0 & 63.2 & 12.5 & 34.8 \\
Without task memory & 34.0 & 51.5 & 54.0 & 65.9 \\
\midrule
\multicolumn{5}{l}{\emph{Component removal, Qwen3-VL-32B}} \\
Full \textsc{Mimir} & \textbf{68.0} & \textbf{73.0} & \textbf{71.5} & \textbf{78.6} \\
Without world memory & 47.0 & 57.2 & 12.5 & 35.0 \\
Without task memory & 30.5 & 46.5 & 46.5 & 60.8 \\
\midrule
\multicolumn{5}{l}{\emph{Reasoning mode, Qwen3-VL-32B}} \\
Instruct & 50.5 & 59.5 & 50.5 & 57.5 \\
Thinking & 48.0 & 55.8 & 52.0 & 61.7 \\
Instruct, \textsc{Mimir} & \textbf{68.0} & \textbf{73.0} & \textbf{71.5} & \textbf{78.6} \\
Thinking, \textsc{Mimir} & 61.5 & 69.1 & 67.5 & 75.6 \\
\bottomrule
\end{tabular}
\caption{Four-task average ablations. The first two blocks remove task memory or world memory from \textsc{Mimir}. The final block compares the Instruct and Thinking variants of Qwen3-VL-32B under the backbone and \textsc{Mimir} settings.}
\label{tab:component_ablation}
\end{table}

The component removals expose a stable division of labor. On EB-Habitat, removing world memory reduces SR from 65.0 to 12.5 with Qwen3-VL-8B and from 71.5 to 12.5 with Qwen3-VL-32B. On EB-ALFRED, removing task memory causes the larger loss, reducing SR by 23.0 points and 37.5 points at the two scales. World memory therefore supplies the object and location evidence needed to instantiate the active goal, whereas task memory preserves ordered progress and execution constraints. Full \textsc{Mimir} is strongest across both scales and environments, showing that the two states are most effective when coupled through grounding.

The reasoning-mode comparison holds model size fixed. Without \textsc{Mimir}, Thinking is slightly stronger on EB-Habitat but weaker on EB-ALFRED. With \textsc{Mimir}, Instruct is higher on all four metrics, reaching an SR of 68.0 and a GC of 73.0 on EB-ALFRED, together with an SR of 71.5 and a GC of 78.6 on EB-Habitat. Thinking still gains 13.5 SR points and 13.3 GC points on EB-ALFRED, and 15.5 SR points and 13.9 GC points on EB-Habitat. Internal reasoning and explicit memory are therefore not redundant. Instead, the reversal suggests that once task progress and world evidence are externalized, the Instruct model can exploit the grounded decision state more directly, while the additional reasoning mode provides no further advantage in this setting.

\subsection{Failure Analysis}

\begin{figure}[t]
    \centering
    \begin{subfigure}[t]{0.48\linewidth}
        \centering
        \includegraphics[width=\linewidth]{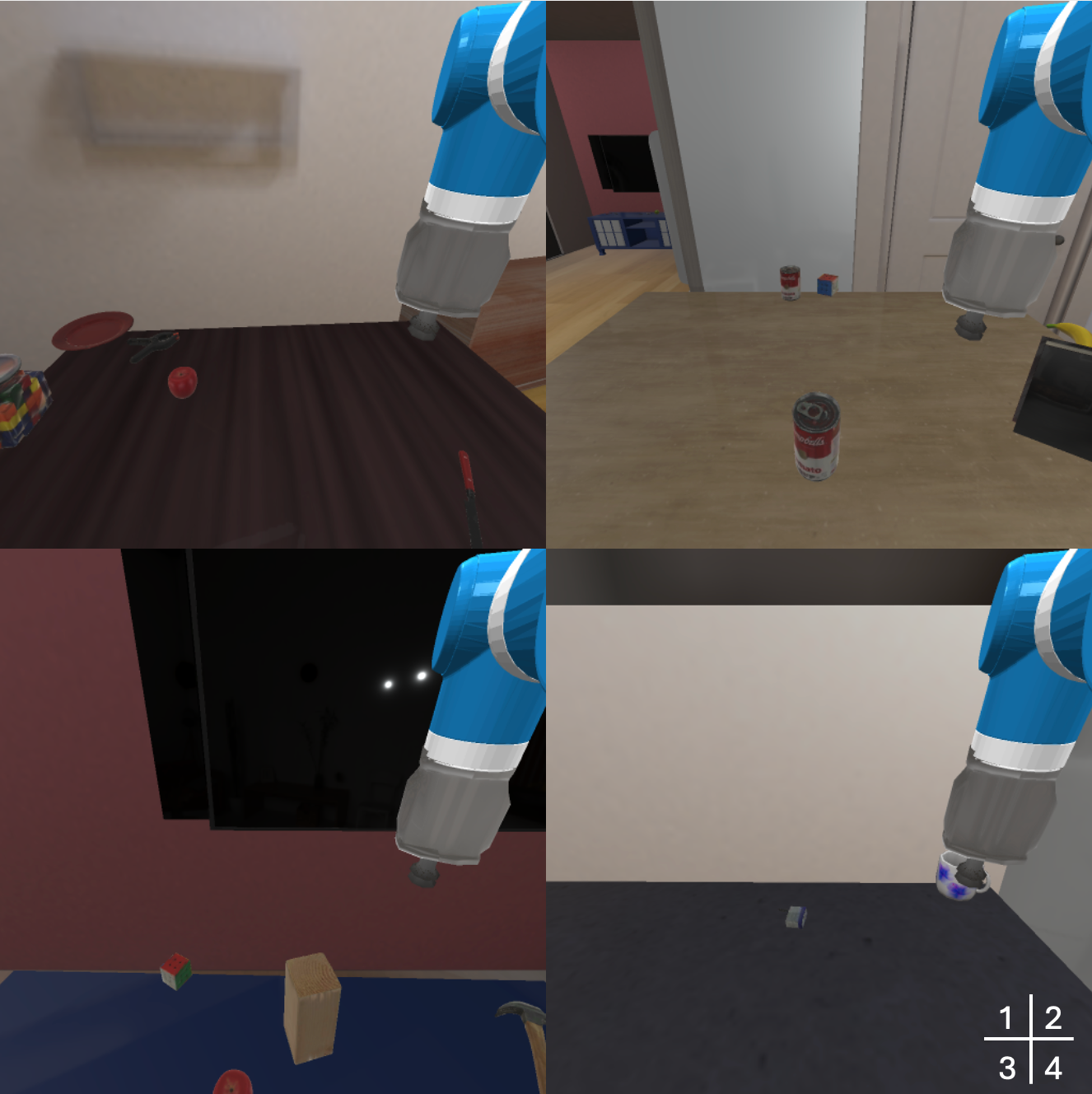}
        \caption{Failures caused by ambiguous target descriptions and insufficient visual evidence.}
        \label{fig:wrong}
    \end{subfigure}
    \hfill
    \begin{subfigure}[t]{0.48\linewidth}
        \centering
        \includegraphics[width=\linewidth]{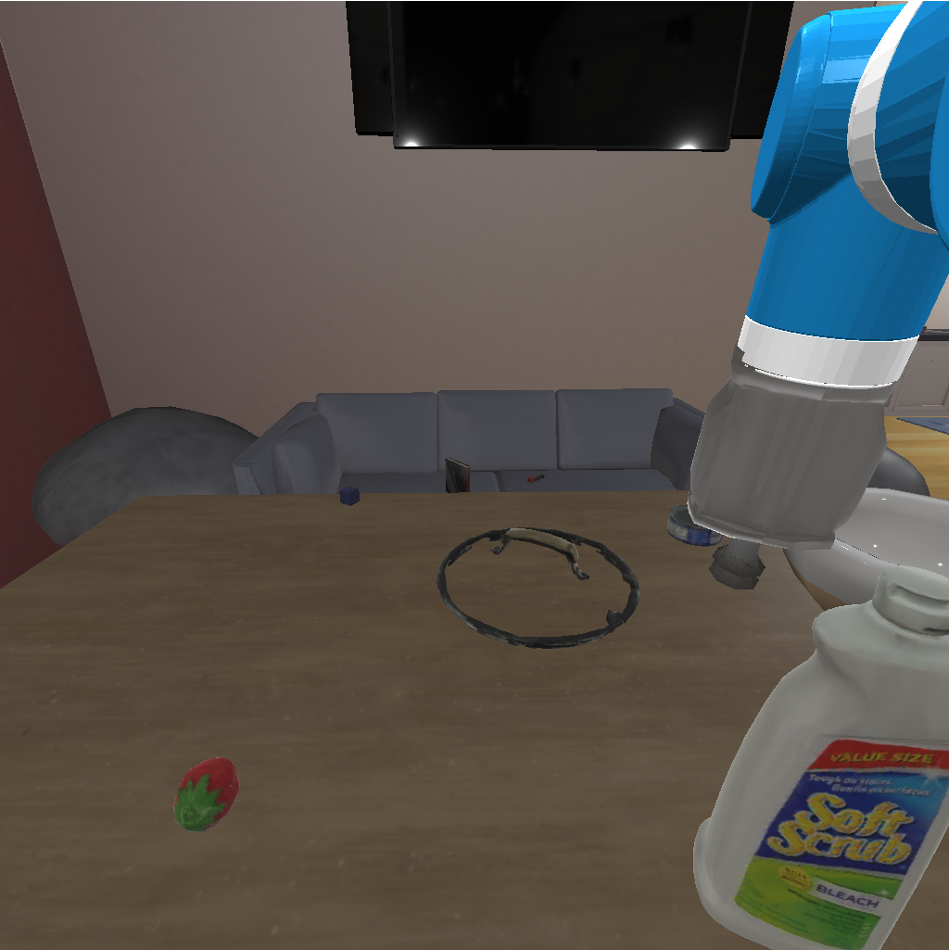}
        \caption{Failures caused by unrealistic environment configurations and severe occlusion.}
        \label{fig:wrong2}
    \end{subfigure}
    \caption{Representative failure cases under ambiguous instructions, insufficient visual evidence, unrealistic environment configurations, and severe occlusion.}
    \label{fig:failure_cases}
\end{figure}

We inspect failed long-horizon episodes and find two recurring information bottlenecks. First, some instructions identify a target only by a coarse category or color, while several scene instances satisfy the same description, as shown in panels~1,2 of Fig.~\ref{fig:wrong}. \textsc{Mimir} can retain and compare these candidates, but the available input does not determine which instance matches the evaluator annotation. The failure therefore occurs at the boundary between language specification and instance-level grounding.

Second, the target may be invisible, or rendered without enough detail for reliable recognition, as shown in panels~3,4 of Fig.~\ref{fig:wrong} and Fig.~\ref{fig:wrong2}. These observations produce either no world memory entry or an entry too vague to support later retrieval. Memory can preserve evidence that has been observed, but it cannot reconstruct evidence that never entered the perceptual stream. These cases delimit the role of \textsc{Mimir}. It organizes and connects available evidence, but it cannot resolve a target that is not identifiable from the agent's inputs.

\section{Conclusion}

We presented \textsc{Mimir}, a neuro-symbolic memory system that
treats long-horizon embodied execution as the coordinated evolution
of world memory for scene belief and task memory for execution
progress. Dynamic grounding connects these states by turning the
active goal and recalled evidence into an actionable decision state
before task-directed execution. This design shifts memory from passive
storage to an online interface for action selection, allowing scene
belief and execution progress to evolve independently while informing
each other. Experiments across diverse backbones and embodied
environments demonstrate the effectiveness of this task--world
interface. Our results suggest that reliable long-horizon memory
depends not simply on retaining more history, but on organizing and
retrieving the state needed for the current goal.

\clearpage
\bibliography{references}

@incollection{baddeley1974working,
  author = {Alan D. Baddeley and Graham Hitch},
  title = {Working Memory},
  booktitle = {Psychology of Learning and Motivation},
  volume = {8},
  pages = {47--89},
  publisher = {Academic Press},
  year = {1974}
}

@incollection{tulving1972episodic,
  author = {Endel Tulving},
  title = {Episodic and Semantic Memory},
  booktitle = {Organization of Memory},
  pages = {381--403},
  publisher = {Academic Press},
  year = {1972}
}

@misc{yao2023react,
  author = {Shunyu Yao and Jeffrey Zhao and Dian Yu and Nan Du and Izhak Shafran and Karthik Narasimhan and Yuan Cao},
  title = {{ReAct}: Synergizing Reasoning and Acting in Language Models},
  year = {2023},
  eprint = {2210.03629},
  archivePrefix = {arXiv},
  note = {ICLR 2023}
}

@misc{shinn2023reflexion,
  author = {Noah Shinn and Federico Cassano and Edward Berman and Ashwin Gopinath and Karthik Narasimhan and Shunyu Yao},
  title = {Reflexion: Language Agents with Verbal Reinforcement Learning},
  year = {2023},
  eprint = {2303.11366},
  archivePrefix = {arXiv}
}

@misc{wang2023voyager,
  author = {Guanzhi Wang and Yuqi Xie and Yunfan Jiang and Ajay Mandlekar and Chaowei Xiao and Yuke Zhu and Linxi Fan and Anima Anandkumar},
  title = {Voyager: An Open-Ended Embodied Agent with Large Language Models},
  year = {2023},
  eprint = {2305.16291},
  archivePrefix = {arXiv}
}

@misc{packer2023memgpt,
  author = {Charles Packer and Sarah Wooders and Kevin Lin and Vivian Fang and Shishir G. Patil and Ion Stoica and Joseph E. Gonzalez},
  title = {{MemGPT}: Towards {LLMs} as Operating Systems},
  year = {2023},
  eprint = {2310.08560},
  archivePrefix = {arXiv}
}

@misc{ahn2022saycan,
  author = {Michael Ahn and Anthony Brohan and Noah Brown and Yevgen Chebotar and Omar Cortes and Byron David and Chelsea Finn and Chuyuan Fu and Keerthana Gopalakrishnan and Karol Hausman and Alex Herzog and Daniel Ho and Jasmine Hsu and Julian Ibarz and Brian Ichter and Alex Irpan and Eric Jang and Rosario Jauregui Ruano and Kyle Jeffrey and Sally Jesmonth and Nikhil J. Joshi and Ryan Julian and Dmitry Kalashnikov and Yuheng Kuang and Kuang-Huei Lee and Sergey Levine and Yao Lu and Linda Luu and Carolina Parada and Peter Pastor and Jornell Quiambao and Kanishka Rao and Jarek Rettinghouse and Diego Reyes and Pierre Sermanet and Nicolas Sievers and Clayton Tan and Alexander Toshev and Vincent Vanhoucke and Fei Xia and Ted Xiao and Peng Xu and Sichun Xu and Mengyuan Yan and Andy Zeng},
  title = {Do As I Can, Not As I Say: Grounding Language in Robotic Affordances},
  year = {2022},
  eprint = {2204.01691},
  archivePrefix = {arXiv}
}

@misc{song2023llmplanner,
  author = {Chan Hee Song and Jiaman Wu and Clayton Washington and Brian M. Sadler and Wei-Lun Chao and Yu Su},
  title = {{LLM-Planner}: Few-Shot Grounded Planning for Embodied Agents with Large Language Models},
  year = {2023},
  eprint = {2212.04088},
  archivePrefix = {arXiv},
  note = {ICCV 2023}
}

@misc{driess2023palme,
  author = {Danny Driess and Fei Xia and Mehdi S. M. Sajjadi and Corey Lynch and Aakanksha Chowdhery and Brian Ichter and Ayzaan Wahid and Jonathan Tompson and Quan Vuong and Tianhe Yu and Wenlong Huang and Yevgen Chebotar and Pierre Sermanet and Daniel Duckworth and Sergey Levine and Vincent Vanhoucke and Karol Hausman and Marc Toussaint and Klaus Greff and Andy Zeng and Igor Mordatch and Pete Florence},
  title = {{PaLM-E}: An Embodied Multimodal Language Model},
  year = {2023},
  eprint = {2303.03378},
  archivePrefix = {arXiv}
}

@misc{chen2024automanual,
  author = {Minghao Chen and Yihang Li and Yanting Yang and Shiyu Yu and Binbin Lin and Xiaofei He},
  title = {{AutoManual}: Constructing Instruction Manuals by {LLM} Agents via Interactive Environmental Learning},
  year = {2024},
  eprint = {2405.16247},
  archivePrefix = {arXiv},
  note = {NeurIPS 2024}
}

@misc{fu2024msiagent,
  author = {Dayuan Fu and Biqing Qi and Yihuai Gao and Che Jiang and Guanting Dong and Bowen Zhou},
  title = {{MSI-Agent}: Incorporating Multi-Scale Insight into Embodied Agents for Superior Planning and Decision-Making},
  year = {2024},
  eprint = {2409.16686},
  archivePrefix = {arXiv},
  note = {EMNLP 2024}
}

@misc{karma2024,
  author = {Zixuan Wang and Bo Yu and Junzhe Zhao and Wenhao Sun and Sai Hou and Shuai Liang and Xing Hu and Yinhe Han and Yiming Gan},
  title = {{KARMA}: Augmenting Embodied {AI} Agents with Long-and-short Term Memory Systems},
  year = {2024},
  eprint = {2409.14908},
  archivePrefix = {arXiv}
}

@misc{lei2025robomemory,
  author = {Mingcong Lei and Honghao Cai and Yuyuan Yang and Yimou Wu and Jinke Ren and Zezhou Cui and Liangchen Tan and Junkun Hong and Gehan Hu and Shuangyu Zhu and Shaohan Jiang and Ge Wang and Junyuan Tan and Zhenglin Wan and Zheng Li and Zhen Li and Shuguang Cui and Yiming Zhao and Yatong Han},
  title = {{RoboMemory}: A Brain-inspired Multi-memory Agentic Framework for Interactive Environmental Learning in Physical Embodied Systems},
  year = {2025},
  eprint = {2508.01415},
  archivePrefix = {arXiv}
}

@article{chen2023llmstate,
  author = {Siwei Chen and Anxing Xiao and David Hsu},
  title = {{LLM-State}: Open World State Representation for Long-Horizon Task Planning with Large Language Model},
  journal = {arXiv preprint arXiv:2311.17406},
  year = {2023}
}

@inproceedings{kim2023capeam,
  author = {Byeonghwi Kim and Jinyeon Kim and Yuyeong Kim and Cheolhong Min and Jonghyun Choi},
  title = {Context-Aware Planning and Environment-Aware Memory for Instruction Following Embodied Agents},
  booktitle = {Proceedings of the IEEE/CVF International Conference on Computer Vision},
  pages = {10936--10946},
  year = {2023}
}

@inproceedings{rana2023sayplan,
  author = {Krishan Rana and Jesse Haviland and Sourav Garg and Jad Abou-Chakra and Ian Reid and Niko S{\"u}nderhauf},
  title = {SayPlan: Grounding Large Language Models Using {3D} Scene Graphs for Scalable Robot Task Planning},
  booktitle = {Proceedings of the 7th Conference on Robot Learning},
  pages = {23--72},
  volume = {229},
  series = {Proceedings of Machine Learning Research},
  year = {2023}
}

@misc{gu2024conceptgraphs,
  author = {Qiao Gu and Alihusein Kuwajerwala and Sacha Morin and Krishna Murthy Jatavallabhula and Bipasha Sen and Aditya Agarwal and Corban Rivera and William Paul and Kirsty Ellis and Rama Chellappa and Chuang Gan and Celso Miguel de Melo and Joshua B. Tenenbaum and Antonio Torralba and Florian Shkurti and Liam Paull},
  title = {{ConceptGraphs}: Open-Vocabulary {3D} Scene Graphs for Perception and Planning},
  year = {2023},
  eprint = {2309.16650},
  archivePrefix = {arXiv}
}

@misc{yang20253dmem,
  author = {Yuncong Yang and Han Yang and Jiachen Zhou and Peihao Chen and Hongxin Zhang and Yilun Du and Chuang Gan},
  title = {{3D-Mem}: {3D} Scene Memory for Embodied Exploration and Reasoning},
  year = {2024},
  eprint = {2411.17735},
  archivePrefix = {arXiv}
}

@misc{brohan2023rt2,
  author = {Anthony Brohan and Noah Brown and Justice Carbajal and Yevgen Chebotar and Xi Chen and Krzysztof Choromanski and Tianli Ding and Danny Driess and Avinava Dubey and Chelsea Finn and Pete Florence and Chuyuan Fu and Montse Gonzalez Arenas and Keerthana Gopalakrishnan and Kehang Han and Karol Hausman and Alexander Herzog and Jasmine Hsu and Brian Ichter and Alex Irpan and Nikhil Joshi and Ryan Julian and Dmitry Kalashnikov and Yuheng Kuang and Isabel Leal and Lisa Lee and Tsang-Wei Edward Lee and Sergey Levine and Yao Lu and Henryk Michalewski and Igor Mordatch and Karl Pertsch and Kanishka Rao and Krista Reymann and Michael Ryoo and Grecia Salazar and Pannag Sanketi and Pierre Sermanet and Jaspiar Singh and Anikait Singh and Radu Soricut and Huong Tran and Vincent Vanhoucke and Quan Vuong and Ayzaan Wahid and Stefan Welker and Paul Wohlhart and Jialin Wu and Fei Xia and Ted Xiao and Peng Xu and Sichun Xu and Tianhe Yu and Brianna Zitkovich},
  title = {{RT-2}: Vision-Language-Action Models Transfer Web Knowledge to Robotic Control},
  year = {2023},
  eprint = {2307.15818},
  archivePrefix = {arXiv}
}

@misc{kim2024openvla,
  author = {Moo Jin Kim and Karl Pertsch and Siddharth Karamcheti and Ted Xiao and Ashwin Balakrishna and Suraj Nair and Rafael Rafailov and Ethan Foster and Grace Lam and Pannag Sanketi and Quan Vuong and Thomas Kollar and Benjamin Burchfiel and Russ Tedrake and Dorsa Sadigh and Sergey Levine and Percy Liang and Chelsea Finn},
  title = {{OpenVLA}: An Open-Source Vision-Language-Action Model},
  year = {2024},
  eprint = {2406.09246},
  archivePrefix = {arXiv}
}

@misc{memoryvla2025,
  author = {Hao Shi and Bin Xie and Yingfei Liu and Lin Sun and Fengrong Liu and Tiancai Wang and Erjin Zhou and Haoqiang Fan and Xiangyu Zhang and Gao Huang},
  title = {{MemoryVLA}: Perceptual-Cognitive Memory in Vision-Language-Action Models for Robotic Manipulation},
  year = {2025},
  eprint = {2508.19236},
  archivePrefix = {arXiv},
  note = {ICLR 2026}
}

@misc{memer2025,
  author = {Ajay Sridhar and Jennifer Pan and Satvik Sharma and Chelsea Finn},
  title = {{MemER}: Scaling Up Memory for Robot Control via Experience Retrieval},
  year = {2025},
  eprint = {2510.20328},
  archivePrefix = {arXiv}
}

@misc{mapvla2025,
  author = {Runhao Li and Wenkai Guo and Zhenyu Wu and Changyuan Wang and Haoyuan Deng and Zhenyu Weng and Yap-Peng Tan and Ziwei Wang},
  title = {{MAP-VLA}: Memory-Augmented Prompting for Vision-Language-Action Model in Robotic Manipulation},
  year = {2025},
  eprint = {2511.09516},
  archivePrefix = {arXiv}
}

@misc{yang2025embodiedbench,
  author = {Rui Yang and Hanyang Chen and Junyu Zhang and Mark Zhao and Cheng Qian and Kangrui Wang and Qineng Wang and Teja Venkat Koripella and Marziyeh Movahedi and Manling Li and Heng Ji and Huan Zhang and Tong Zhang},
  title = {{EmbodiedBench}: Comprehensive Benchmarking Multi-modal Large Language Models for Vision-Driven Embodied Agents},
  year = {2025},
  eprint = {2502.09560},
  archivePrefix = {arXiv}
}

@incollection{atkinson1968human,
  author = {Richard C. Atkinson and Richard M. Shiffrin},
  title = {Human Memory: A Proposed System and Its Control Processes},
  booktitle = {Psychology of Learning and Motivation},
  volume = {2},
  pages = {89--195},
  publisher = {Academic Press},
  year = {1968}
}

@misc{chhikara2025mem0,
  author = {Prateek Chhikara and Dev Khant and Saket Aryan and Taranjeet Singh and Deshraj Yadav},
  title = {{Mem0}: Building Production-Ready {AI} Agents with Scalable Long-Term Memory},
  year = {2025},
  eprint = {2504.19413},
  archivePrefix = {arXiv}
}

@misc{zhang2023coela,
  author = {Hongxin Zhang and Weihua Du and Jiaming Shan and Qinhong Zhou and Yilun Du and Joshua B. Tenenbaum and Tianmin Shu and Chuang Gan},
  title = {Building Cooperative Embodied Agents Modularly with Large Language Models},
  year = {2023},
  eprint = {2307.02485},
  archivePrefix = {arXiv}
}

@misc{lei2025clea,
  author = {Mingcong Lei and Ge Wang and Yiming Zhao and Zhixin Mai and Qing Zhao and Yao Guo and Zhen Li and Shuguang Cui and Yatong Han and Jinke Ren},
  title = {{CLEA}: Closed-Loop Embodied Agent for Enhancing Task Execution in Dynamic Environments},
  year = {2025},
  eprint = {2503.00729},
  archivePrefix = {arXiv}
}

@misc{glocker2025llm,
  author = {Marc Glocker and Peter H{\"o}nig and Matthias Hirschmanner and Markus Vincze},
  title = {{LLM}-Empowered Embodied Agent for Memory-Augmented Task Planning in Household Robotics},
  year = {2025},
  eprint = {2504.21716},
  archivePrefix = {arXiv}
}

@misc{choi2025nesyc,
  author = {Wonje Choi and Jinwoo Park and Sanghyun Ahn and Daehee Lee and Honguk Woo},
  title = {{NeSyC}: A Neuro-symbolic Continual Learner For Complex Embodied Tasks In Open Domains},
  year = {2025},
  eprint = {2503.00870},
  archivePrefix = {arXiv},
  note = {ICLR 2025}
}

@misc{ma2026brainmem,
  author = {Xiaoyu Ma and Lianyu Hu and Wenbing Tang and Zixuan Hu and Zeqin Liao and Zhizhen Wu and Yang Liu},
  title = {{BrainMem}: Brain-Inspired Evolving Memory for Embodied Agent Task Planning},
  year = {2026},
  eprint = {2604.16331},
  archivePrefix = {arXiv}
}

@misc{memcompiler2026,
  author = {Xin Ding and Xinrui Wang and Yifan Yang and Hao Wu and Shiqi Jiang and Qianxi Zhang and Liang Mi and Hanxin Zhu and Kun Li and Yunxin Liu and Zhibo Chen and Ting Cao},
  title = {{MemCompiler}: Compile, Don't Inject -- State-Conditioned Memory for Embodied Agents},
  year = {2026},
  eprint = {2605.07594},
  archivePrefix = {arXiv}
}

@misc{yang2026keep,
  author = {Zebin Yang and Tong Xie and Baotong Lu and Shaoshan Liu and Bo Yu and Meng Li},
  title = {{KEEP}: A {KV}-Cache-Centric Memory Management System for Efficient Embodied Planning},
  year = {2026},
  eprint = {2602.23592},
  archivePrefix = {arXiv}
}

@misc{huang2022vlmaps,
  author = {Chenguang Huang and Oier Mees and Andy Zeng and Wolfram Burgard},
  title = {Visual Language Maps for Robot Navigation},
  year = {2022},
  eprint = {2210.05714},
  archivePrefix = {arXiv}
}

@misc{jatavallabhula2023conceptfusion,
  author = {Krishna Murthy Jatavallabhula and Alihusein Kuwajerwala and Qiao Gu and Mohd Omama and Tao Chen and Alaa Maalouf and Shuang Li and Ganesh Iyer and Soroush Saryazdi and Nikhil Keetha and Ayush Tewari and Joshua B. Tenenbaum and Celso Miguel de Melo and Madhava Krishna and Liam Paull and Florian Shkurti and Antonio Torralba},
  title = {{ConceptFusion}: Open-set Multimodal {3D} Mapping},
  year = {2023},
  eprint = {2302.07241},
  archivePrefix = {arXiv}
}

@misc{chang2023goat,
  author = {Matthew Chang and Theophile Gervet and Mukul Khanna and Sriram Yenamandra and Dhruv Shah and So Yeon Min and Kavit Shah and Chris Paxton and Saurabh Gupta and Dhruv Batra and Roozbeh Mottaghi and Jitendra Malik and Devendra Singh Chaplot},
  title = {{GOAT}: GO to Any Thing},
  year = {2023},
  eprint = {2311.06430},
  archivePrefix = {arXiv}
}

@misc{loo2025openscenegraphs,
  author = {Joel Loo and Zhanxin Wu and David Hsu},
  title = {Open Scene Graphs for Open-World Object-Goal Navigation},
  year = {2025},
  eprint = {2508.04678},
  archivePrefix = {arXiv}
}

@misc{3dllmmem2025,
  author = {Wenbo Hu and Yining Hong and Yanjun Wang and Leison Gao and Zibu Wei and Xingcheng Yao and Nanyun Peng and Yonatan Bitton and Idan Szpektor and Kai-Wei Chang},
  title = {{3DLLM-Mem}: Long-Term Spatial-Temporal Memory for Embodied {3D} Large Language Model},
  year = {2025},
  eprint = {2505.22657},
  archivePrefix = {arXiv}
}

@misc{metamemory2025,
  author = {Yufan Mao and Hanjing Ye and Wenlong Dong and Chengjie Zhang and Hong Zhang},
  title = {{Meta-Memory}: Retrieving and Integrating Semantic-Spatial Memories for Robot Spatial Reasoning},
  year = {2025},
  eprint = {2509.20754},
  archivePrefix = {arXiv}
}

@misc{zhang2025nava3,
  author = {Lingfeng Zhang and Xiaoshuai Hao and Yingbo Tang and Haoxiang Fu and Xinyu Zheng and Pengwei Wang and Zhongyuan Wang and Wenbo Ding and Shanghang Zhang},
  title = {{$NavA^3$}: Understanding Any Instruction, Navigating Anywhere, Finding Anything},
  year = {2025},
  eprint = {2508.04598},
  archivePrefix = {arXiv}
}

@misc{oneill2023openxembodiment,
  author = {{Open X-Embodiment Collaboration} and Abby O'Neill and Abdul Rehman and Abhinav Gupta and Abhiram Maddukuri and Abhishek Gupta and others},
  title = {Open {X}-Embodiment: Robotic Learning Datasets and {RT-X} Models},
  year = {2023},
  eprint = {2310.08864},
  archivePrefix = {arXiv}
}

@misc{black2024pi0,
  author = {Kevin Black and Noah Brown and Danny Driess and Adnan Esmail and Michael Equi and Chelsea Finn and Niccolo Fusai and Lachy Groom and Karol Hausman and Brian Ichter and Szymon Jakubczak and Tim Jones and Liyiming Ke and Sergey Levine and Adrian Li-Bell and Mohith Mothukuri and Suraj Nair and Karl Pertsch and Lucy Xiaoyang Shi and James Tanner and Quan Vuong and Anna Walling and Haohuan Wang and Ury Zhilinsky},
  title = {{$\pi_0$}: A Vision-Language-Action Flow Model for General Robot Control},
  year = {2024},
  eprint = {2410.24164},
  archivePrefix = {arXiv}
}

@misc{echovla2025,
  author = {Min Lin and Xiwen Liang and Bingqian Lin and {Liu Jingzhi} and Zijian Jiao and Kehan Li and Yu Sun and Weijia Liufu and Yuhan Ma and Yuecheng Liu and Shen Zhao and Yuzheng Zhuang and Xiaodan Liang},
  title = {{EchoVLA}: Synergistic Declarative Memory for {VLA}-Driven Mobile Manipulation},
  year = {2025},
  eprint = {2511.18112},
  archivePrefix = {arXiv}
}

@misc{chameleon2026,
  author = {Xinying Guo and Chenxi Jiang and Hyun Bin Kim and Ying Sun and Yang Xiao and Yuhang Han and Jianfei Yang},
  title = {{Chameleon}: Episodic Memory for Long-Horizon Robotic Manipulation},
  year = {2026},
  eprint = {2603.24576},
  archivePrefix = {arXiv}
}

@misc{mem2026,
  author = {Marcel Torne and Karl Pertsch and Homer Walke and Kyle Vedder and Suraj Nair and Brian Ichter and Allen Z. Ren and Haohuan Wang and Jiaming Tang and Kyle Stachowicz and Karan Dhabalia and Michael Equi and Quan Vuong and Jost Tobias Springenberg and Sergey Levine and Chelsea Finn and Danny Driess},
  title = {{MEM}: Multi-Scale Embodied Memory for Vision Language Action Models},
  year = {2026},
  eprint = {2603.03596},
  archivePrefix = {arXiv}
}

@misc{rememvla2026,
  author = {Hang Li and Fengyi Shen and Dong Chen and Liudi Yang and Xudong Wang and Jinkui Shi and Zhenshan Bing and Ziyuan Liu and Alois Knoll},
  title = {{ReMem-VLA}: Empowering Vision-Language-Action Model with Memory via Dual-Level Recurrent Queries},
  year = {2026},
  eprint = {2603.12942},
  archivePrefix = {arXiv}
}

@misc{helm2026,
  author = {Zijian Zeng and Fei Ding and Huiming Yang and Xianwei Li},
  title = {{HELM}: Harness-Enhanced Long-horizon Memory for Vision-Language-Action Manipulation},
  year = {2026},
  eprint = {2604.18791},
  archivePrefix = {arXiv}
}

@misc{vpwem2026,
  author = {Yuheng Lei and Zhixuan Liang and Hongyuan Zhang and Ping Luo},
  title = {{VPWEM}: Non-Markovian Visuomotor Policy with Working and Episodic Memory},
  year = {2026},
  eprint = {2603.04910},
  archivePrefix = {arXiv}
}

@misc{physmem2026,
  author = {Haoyang Li and Yang You and Hao Su and Leonidas Guibas},
  title = {{PhysMem}: Scaling Test-Time Memory for Embodied Physical Reasoning},
  year = {2026},
  eprint = {2602.20323},
  archivePrefix = {arXiv}
}

@misc{tan2024cradle,
  author        = {Weihao Tan and Wentao Zhang and Xinrun Xu and Haochong Xia and Ziluo Ding and Boyu Li and Bohan Zhou and Junpeng Yue and Jiechuan Jiang and Yewen Li and Ruyi An and Molei Qin and Chuqiao Zong and Longtao Zheng and Yujie Wu and Xiaoqiang Chai and Yifei Bi and Tianbao Xie and Pengjie Gu and Xiyun Li and Ceyao Zhang and Long Tian and Chaojie Wang and Xinrun Wang and B{\"o}rje F. Karlsson and Bo An and Shuicheng Yan and Zongqing Lu},
  title         = {Cradle: Empowering Foundation Agents Towards General Computer Control},
  year          = {2024},
  eprint        = {2403.03186},
  archivePrefix = {arXiv},
  primaryClass  = {cs.AI},
  doi           = {10.48550/arXiv.2403.03186},
  url           = {https://arxiv.org/abs/2403.03186}
}

@misc{tan2025roboos,
  author        = {Huajie Tan and Xiaoshuai Hao and Cheng Chi and Minglan Lin and Yaoxu Lyu and Mingyu Cao and Dong Liang and Zhuo Chen and Mengsi Lyu and Cheng Peng and Chenrui He and Yulong Ao and Yonghua Lin and Pengwei Wang and Zhongyuan Wang and Shanghang Zhang},
  title         = {{RoboOS}: A Hierarchical Embodied Framework for Cross-Embodiment and Multi-Agent Collaboration},
  year          = {2025},
  eprint        = {2505.03673},
  archivePrefix = {arXiv},
  primaryClass  = {cs.RO},
  doi           = {10.48550/arXiv.2505.03673},
  url           = {https://arxiv.org/abs/2505.03673}
}




\end{document}